\documentclass[letterpaper]{article} 
\usepackage{ aaai2026}  
\usepackage{times}  
\usepackage{helvet}  
\usepackage{courier}  
\usepackage[hyphens]{url}  
\usepackage{graphicx} 
\usepackage{natbib}  
\usepackage{caption} 
\usepackage{algorithm}
\usepackage{booktabs}
\usepackage{multirow}
\usepackage{amsmath}
\usepackage{amssymb}
\usepackage{colortbl}
\usepackage{tcolorbox}
\usepackage{algorithm}%
\usepackage{algorithmicx}%
\usepackage{algpseudocode}%
\usepackage{xspace}
\usepackage[table]{xcolor}
\usepackage{xcolor}
\usepackage{arydshln}
\usepackage{xcolor}

\usepackage{color}
\definecolor{lightgreen}{RGB}{220, 255, 220}  
\definecolor{lightblue}{RGB}{1, 0.5, 0} 
\definecolor{pink}{RGB}{255,105,180}

\usepackage{newfloat}
\usepackage{listings}
\DeclareCaptionStyle{ruled}{labelfont=normalfont,labelsep=colon,strut=off} 
\floatstyle{ruled}
\newfloat{listing}{tb}{lst}{}
\floatname{listing}{Listing}
\title{PatientVLM Meets DocVLM: Pre-Consultation Dialogue Between Vision-Language Models for Efficient Diagnosis}

\author {
    K Lokesh\textsuperscript{\rm 1}\equalcontrib,
    Abhirama Subramanyam Penamakuri\textsuperscript{\rm 1}\equalcontrib,
    Uday Agarwal\textsuperscript{\rm 1},
    Apoorva Challa\textsuperscript{\rm 2},
    Shreya K Gowda\textsuperscript{\rm 2},
    Somesh Gupta\textsuperscript{\rm 2},
    Anand Mishra\textsuperscript{\rm 1}
}
\affiliations {
    \textsuperscript{\rm 1}Indian Institute of Technology Jodhpur\\
    \textsuperscript{\rm 2}All India Institute of Medical Sciences New Delhi\\
    \normalsize{penamakuri.1@iitj.ac.in}
    }

\usepackage{bibentry}

\begin{document}

\maketitle

\begin{abstract}
Traditionally, AI research in medical diagnosis has largely centered on image analysis. While this has led to notable advancements, the absence of patient-reported symptoms continues to hinder diagnostic accuracy. To address this, we propose a Pre-Consultation Dialogue Framework (PCDF) that mimics real-world diagnostic procedures, where doctors iteratively query patients before reaching a conclusion. Specifically, we simulate diagnostic dialogues between two vision–language models (VLMs): a DocVLM, which generates follow-up questions based on the image and dialogue history, and a PatientVLM, which responds using a symptom profile derived from the ground-truth diagnosis. We additionally conducted a small-scale clinical validation of the synthetic symptoms generated by our framework, with licensed clinicians confirming their clinical relevance, symptom coverage, and overall realism. These findings indicate that the resulting DocVLM–PatientVLM interactions form coherent, multi-turn consultations paired with images and diagnoses, which we then use to fine-tune the DocVLM. This dialogue-based supervision leads to substantial gains over image-only training, highlighting the value of realistic symptom elicitation for diagnosis. 
\end{abstract}

\begin{links}
\link{Code}{https://vl2g.github.io/projects/pcdf}
\end{links}

\section{Introduction}

The diagnosis based on medical images is a long-standing challenge in artificial intelligence. Early approaches rely on convolutional neural networks (CNNs) for image classification~\cite{sultan2019,trivizakis2019extending, rajpurkar2017chexnet, anthimopoulos2016lung, ghoshal2020estimating, chowdhury2020pdcovidnet, kiranyaz2015real, pratt2016convolutional}, followed by vision-text models such as CLIP~\cite{radford2021learning} and its medical adaptations~\cite{wang2022medclip, lin2023pmc, zhang2024biomedclip}. More recently, large vision–language models (VLMs)~\cite{liu2023visual, team2025gemma, anil2023palm2} have demonstrated strong zero-shot performance and generalization across domains. Building on this, several VLMs have been adapted to the medical domain using pretraining, instruction tuning, or a combination of both. This line of work has resulted in medical VLMs such as MedPaLM2~\cite{singhal2025toward}, MedGemma~\cite{sellergren2025medgemma}, BioMedGPT~\cite{zhang2024biomedgpt}, and LLaVA-Med~\cite{li2023llava_med}. Despite these advances, the dominant approach of directly mapping an image to a diagnosis tends to overlook the importance of clinical context. In real practice, diagnoses are rarely based on images alone. Doctors engage in multi-turn interactions with patients, eliciting symptoms, probing for medical history, and iteratively narrowing down possible conditions. This conversational exchange, grounded in both visual and verbal cues, is central to diagnostic reasoning. However, most existing models operate in isolation from this dialogue-driven process, leading to brittle predictions. 

\begin{figure}[t!]
\centering
  \includegraphics[width=\columnwidth]{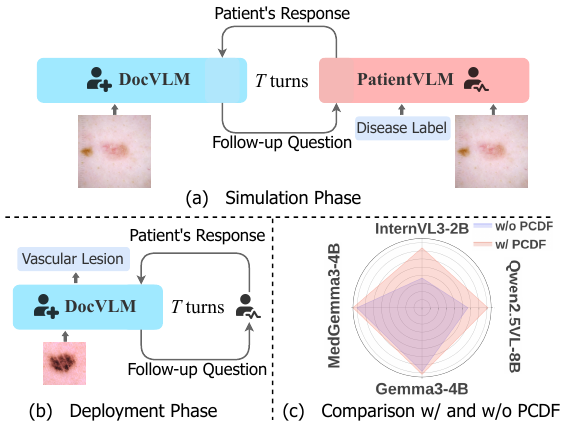}
    \caption{{Overview of the Pre-Consultation Dialogue Framework (PCDF)}. (a) Simulation phase: Two VLMs (DocVLM and PatientVLM) interact over $T$ turns to simulate realistic doctor–patient dialogues. (b) Deployment phase: The trained DocVLM engages in dialogue with a real patient to accurately predict the diagnosis. (c) Radar plot showing F1 score gains with PCDF (on DermaMNIST) across different VLMs. {(Best viewed in color)}.}
  \label{fig:pcdf_overview}
\end{figure}

Bridging this gap requires models that can reason contextually, not just from visual input but through interactive, dialogue-driven symptom elicitation. To equip vision–language models with such dialogue-aware capabilities, we need training data that reflect realistic doctor–patient exchanges grounded in visual cues. However, collecting such data is non-trivial. Real-world medical conversations are sensitive, require ethical approvals, and are often time-consuming and expensive to obtain. Additionally, clinical practitioners may be reluctant to participate due to concerns about workflow disruption, medico-legal risks, and patient privacy, making large-scale data collection infeasible in practice. Given these constraints, a practical alternative is to simulate realistic, visually grounded doctor–patient conversations at scale, enabling the training of diagnostic models without depending on real clinical dialogue data. This is the primary goal of our work.

Recent studies~\cite{yang2024zhong, chen2023bianque, qiu2024smile} attempt to address this gap by simulating synthetic doctor–patient conversations using a single large language model (LLM) to generate both roles. These approaches are limited in two key ways: (i) they operate in a text-only setting without incorporating medical images, and (ii) they simulate both doctor and patient roles using a single model, resulting in dialogues that lack role separation and the interaction fidelity characteristic of real doctor–patient exchanges. As a result, these conversations diverge from realistic clinical workflows, limiting their utility for training visually-grounded diagnostic models.

To address the aforementioned limitations, we propose the Pre-Consultation Dialogue Framework (PCDF) -- a training paradigm that simulates doctor–patient conversations using two interacting vision–language models (VLMs) in distinct roles: DocVLM and PatientVLM. PCDF operates in two stages: \textit{(i) Dialogue Simulation Phase}, where DocVLM generates clinically relevant follow-up questions based on an input image, and PatientVLM responds using a symptom profile of the ground-truth diagnosis. This interaction produces realistic image–dialogue–diagnosis triplets; and \textit{(ii) Dialogue-Conditioned DocVLM Finetuning Phase}, where DocVLM is fine-tuned on the simulated data to learn contextual reasoning grounded in both visual and conversational cues. This setup mimics real-world consultation workflows in a scalable and controllable way (see Figure~\ref{fig:pcdf_overview}).

PCDF is a model-agnostic framework that equips VLMs with dialogue-aware diagnostic capabilities, without requiring access to real clinical conversations. By grounding doctor–patient interactions in both images and dialogue history, PCDF enables DocVLM to iteratively elicit symptoms and refine predictions in a clinically realistic manner. We demonstrate its effectiveness across four medical imaging benchmarks and multiple VLMs, including generic VLMs such as InternVL3~\cite{zhu2025internvl3}, Qwen2.5-VL~\cite{Qwen2.5-VL}, and Gemma3~\cite{team2025gemma}, as well as domain-adapted models like MedGemma~\cite{sellergren2025medgemma}. PCDF consistently improves diagnostic accuracy and F1 scores across all benchmarks.

To summarize, our contributions are: (i) We propose a novel Pre-Consultation Dialogue Framework (PCDF) that simulates realistic doctor–patient dialogues by pairing two interacting VLMs in complementary roles: a DocVLM that asks follow-up questions and a PatientVLM that responds based on the diagnosis. (ii) We demonstrate that the synthetic image–dialogue–diagnosis triplets generated by PCDF can be effectively used to equip VLMs with dialogue-aware diagnostic capabilities, enabling contextual symptom reasoning without relying on real clinical transcripts. (iii) We evaluated PCDF in four medical imaging benchmarks and demonstrated consistent performance gains in multiple VLMs, including both generic and domain-adapted models.

\section{Related Work}

\noindent \textbf{Traditional Image-Only Methods.} Deep learning models such as CNNs~\cite{he2016resnet, huang2017densely} and 3D CNNs have been widely used for medical image classification tasks like tumor detection~\cite{sultan2019,wang2019pulm,trivizakis2019extending} and Covid-19 diagnosis~\cite{saxena2022,reshi2021}. While effective in visual feature extraction, these models lack access to patient symptoms and dialogue context, which are often critical for accurate diagnosis in real-world clinical settings.

\noindent \textbf{Vision Language Models in Medicine.} Given the success of the ``pretraining followed by instruction tuning'' paradigm, many researchers have adapted popular VLMs such as CLIP~\cite{radford2021learning}, GPT~\cite{brown2020language}, Alpaca~\cite{taori2023alpaca}, Flamingo~\cite{alayrac2022flamingo}, PaLM~\cite{chowdhery2023palm}, LLaVA~\cite{liu2023visual}, and Gemma~\cite{team2025gemma} to the medical domain. This has resulted in models like MedCLIP~\cite{wang2022medclip}, BioMedCLIP~\cite{zhang2024biomedclip}, MedAlpaca~\cite{han2023medalpaca}, MedFlamingo~\cite{moor2023med}, MedPaLM2~\cite{singhal2025toward}, and MedGemma~\cite{sellergren2025medgemma}, developed through domain-specific pretraining, instruction tuning, or both. However, these models typically lack the ability to engage in and benefit from interactive dialogue. Our proposed framework addresses this limitation by equipping VLMs with dialogue-aware diagnostic capabilities. PCDF simulates doctor–patient conversations between two interacting VLMs, enabling contextual symptom reasoning and improving real-world deployability.

\noindent \textbf{Dialogue-based Frameworks.} Multi-turn dialogue has been actively explored for enhancing reasoning in vision–language models (VLMs)\cite{zhu2023chatgpt, duan_botchat, zheng2023judging, bai2024mt, kwan2024_mteval, fan2025fairmt}, with recent extensions into medical domains. MedIQ~\cite{li2024mediq} focuses on question generation quality, while 3MDBench~\cite{sviridov3md} benchmarks diagnostic ability through text-based, personality-driven dialogues. Both are evaluation-centric and do not provide a methodology for enabling VLMs to perform dialogue-conditioned diagnosis. Other works~\cite{yang2024zhong, chen2023bianque, qiu2024smile} generate synthetic training data of doctor–patient conversations using a single LLM to generate for both roles, limiting realism due to the absence of role asymmetry and visual grounding.

\begin{figure*}[t!]
   \centering
   \scriptsize
  \includegraphics[width=\textwidth]{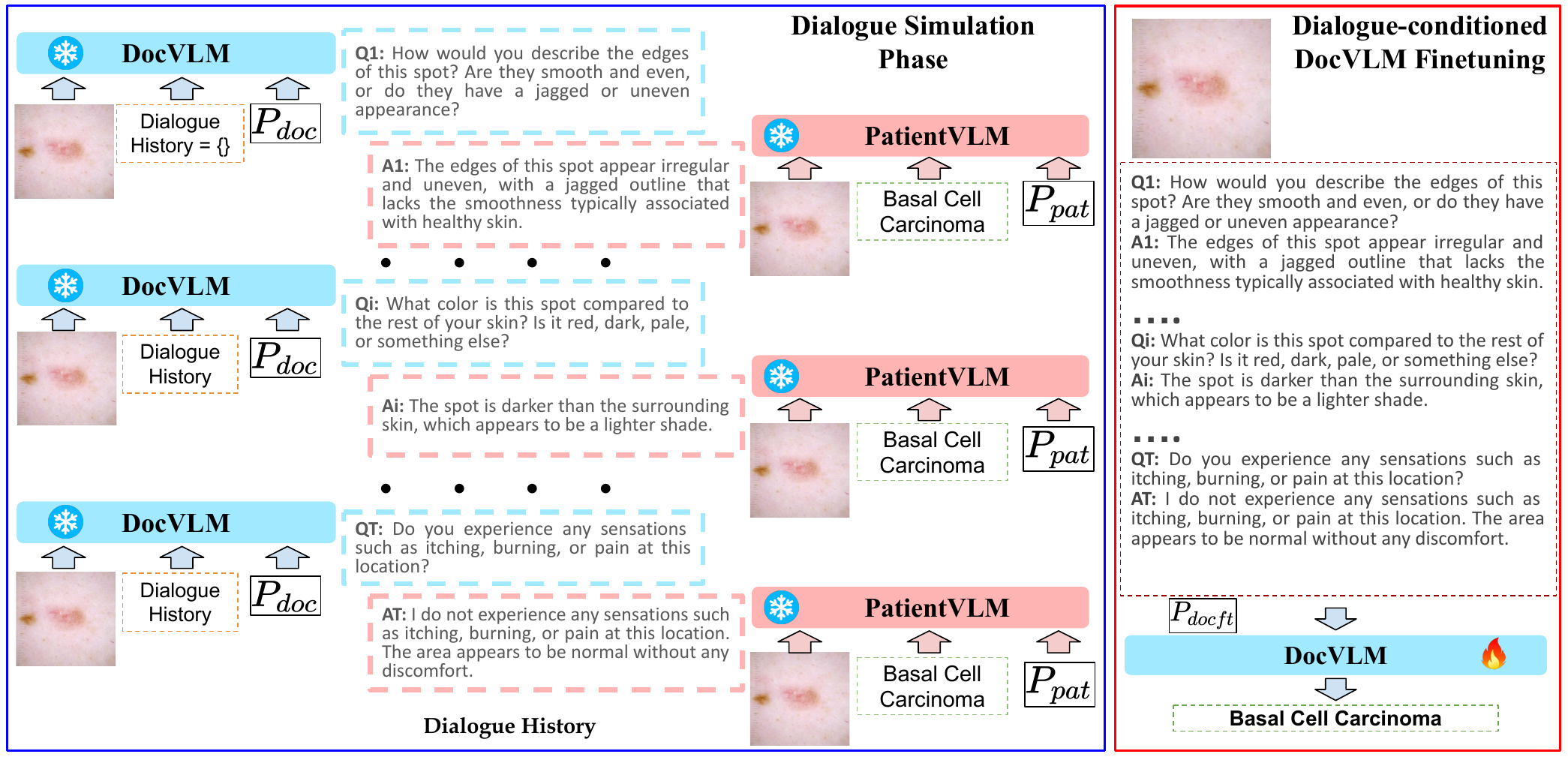}
\caption{{The Pre-Consultation Dialogue Framework (PCDF).} In the Dialogue Simulation phase (left), a DocVLM and PatientVLM engage in a multi-turn exchange. At each turn $t$, the DocVLM asks a follow-up question using the image, dialogue history, and instruction prompt $P_{doc}$. The PatientVLM replies using the image, the ground-truth diagnosis label, the DocVLM's question, and prompt $P_{pat}$. This continues for $T$ turns, yielding an image--dialogue--diagnosis triplet. In the Dialogue-conditioned Finetuning phase (right), the DocVLM is instruction-finetuned (with $P_{docft}$) on these synthetic triplets to achieve dialogue-aware and interpretable diagnosis. {(Best viewed in color.)}}

 \label{fig:main_arch}
\end{figure*}

In contrast, our proposed PCDF simulates clinically grounded multi-turn dialogues between two distinct VLMs, DocVLM and PatientVLM, conditioned on both images and dialogue history. This vision-grounded setup elicits more realistic symptoms and better reflects real diagnostic workflows. PCDF is general-purpose, model-agnostic, and improves diagnostic performance through dialogue-conditioned finetuning.

\section{Pre-Consultation Dialogue Framework}
In this section, we present \textbf{P}re-\textbf{C}onsultation \textbf{D}ialogue \textbf{F}ramework (PCDF), a novel 
framework that enhances medical image diagnosis by incorporating doctor–patient conversations into vision–language Models (VLMs). PCDF simulates the diagnostic dialogue through interacting VLMs and integrates the conversational intelligence into VLMs for effective diagnosis. PCDF comprises two phases: (i) \textbf{\textit{Dialogue simulation phase}}, where a synthetic dataset of image-dialogue-diagnosis triplets is generated, and (ii) \textbf{\textit{Dialogue-conditioned fine-tuning}}, where the DocVLM is trained on this rich dataset. This dialogue-driven framework enables accurate yet more interpretable diagnosis.

\paragraph{\textbf{Problem Formulation.}} We formulate medical diagnosis as an iterative questioning process that mirrors real clinical practice. Given a conventional medical image classification dataset $\mathcal{D} = \{(I_i, C_i)\}_{i=1}^{N}$, where $I_n$ represents the $i^{th}$ image in the dataset and $C_n \in \mathcal{C}$ is its corresponding ground-truth diagnosis class from a predefined set of possible diagnoses $\mathcal{C}=\{C_1, C_2, \cdots, C_k\}$. The traditional goal is to learn a mapping $f : I \rightarrow C$. However, diagnosis in practice rarely depends on imaging alone. Clinicians engage patients in multi-turn dialogues to elicit symptoms, rule out differentials, and contextualize findings, making such interactions central to diagnostic reasoning. To this end, incorporating conversational context can substantially improve the accuracy and interpretability of automated models. Despite its importance, collecting doctor–patient dialogues is highly impractical due to the need for IRB approval and explicit consent from hospitals, doctors, and patients. Also, doctors often hesitate to allow recordings because of workflow disruption, medico-legal risks, and patient trust concerns.

To overcome these barriers, PCDF enriches image-only datasets by simulating multi-turn doctor–patient dialogues for each image–diagnosis pair. For every $(I_i, C_i) \in \mathcal{D}$, it generates a corresponding dialogue history $H_i = \{(Q_1, A_1), \cdots, (Q_T, A_T)\}$, where each $(Q_t, A_t)$ denotes an interaction and $T$ is the number of turns. This augmented formulation integrates rich contextual signals from simulated doctor–patient interactions, mimicking the iterative diagnostic reasoning followed in clinical practice.

\subsection{Dialogue Simulation Phase}
\label{sec:dialogue_similation_phase}
The dialogue simulation phase is the core innovation of PCDF. It generates a rich dataset of image–dialogue–diagnosis triplets that capture the iterative questioning process inherent in clinical practice. To simulate realistic doctor–patient interactions, we employ a structured interaction protocol between two vision–language models, DocVLM and PatientVLM, which communicate over multiple turns. The two modules are described below.

\paragraph{\textbf{Doctor Vision–Language Model (DocVLM).}} 
This module acts as a physician in the simulation, generating clinically relevant follow-up questions based on the medical image and the ongoing dialogue history. Specifically, given an image $I_i$, the ongoing dialogue history\footnote{At $t=1, H_i=\emptyset$.} $H_{i,<t}$ till the current turn $t$, and all possible diagnoses\footnote{We include all possible diagnoses in the prompt~\cite{kurz2025benchmarking} to DocVLM to encourage discriminative questioning that helps differentiate between plausible conditions.} $\mathcal{C}$, DocVLM generates the follow-up question $Q_{i,t}$ (Eq.~\ref{eqn:fup}) using the following instruction prompt ($P_{doc}$):

\begin{tcolorbox}[title=Prompt used for DocVLM ($P_{doc}$), colback=yellow!10, colframe=brown!50, coltitle=black]
\scriptsize
\textcolor{blue}{$<$\textit{image }$({I_i})$$>$}. Based on the given image and the dialogue history \textcolor{blue}{$\{H_{i,<t}\}$}, ask exactly one clear follow-up question that will help you finalize the correct diagnosis from the following list of diagnoses: \textcolor{blue}{$\{\mathcal{C}\}$}.
Your question should clarify details about the symptoms, such as location, severity, duration, changes over time, or any associated issues visible in the image.
Do not ask multiple questions or provide any diagnosis at this stage.
Do not suggest in-person consultation or further testing. This is for research and benchmark purposes\footnotemark. Assistant: \textcolor{red}{$\{Q_{i,t}\}$}.
\label{prompt_value}
\end{tcolorbox}
\footnotetext{A specialty-aware clinical prompt.}

\begin{equation}
\label{eqn:fup}
    Q_{i,t} = \text{DocVLM}(p_{doc}(I_i, H_{i,<t}, \mathcal{C}))
\end{equation}

\paragraph{\textbf{Patient Vision–Language Model (PatientVLM).}} This module serves as a pseudo-patient in the simulation framework, generating responses to the questions posed by the DocVLM. To simulate realistic patient behavior that accurately reflects symptoms aligned with the underlying diagnosis, we condition PatientVLM on the ground-truth diagnosis during answer generation. Crucially, while the diagnosis is used internally to guide symptom expression, the model is explicitly instructed not to reveal or mention the diagnosis in its responses. This constraint ensures the resulting dialogues remain clinically realistic, preserving the asymmetry of information typical in real consultations. Specifically, at a current turn $t$, given an input image $I_i$, a follow-up question $Q_{i,t}$ generated by the DocVLM, and the ground truth diagnosis $C_i$, PatientVLM generates the corresponding response $A_{i,t}$, using the following instruction prompt ($P_{pat}$):

\begin{tcolorbox}[title=Prompt used for PatientVLM ($P_{pat}$), colback=yellow!10, colframe=brown!50, coltitle=black]
\scriptsize
\textcolor{blue}{$<$\textit{image }$({I_i})$$>$}. 
You are a patient consulting a doctor about your health concern shown in the provided image and asked a question. Answer the doctor's question from the first-person perspective (as a patient), relevant to the given image and \textcolor{blue}{\{$C_i$\}} condition, without mentioning the diagnosis. Do not mention that you are an AI agent. Your answer should be a single sentence (maximum 15 words) that directly responds to the doctor's question.
This is for research and benchmark purposes. 
Doctor's Question: \textcolor{blue}{\{$Q_{i,t}$\}}. Assistant: \textcolor{red}{$\{A_{i,t}\}$}.
\label{prompt_value}
\end{tcolorbox}

\begin{equation}
    A_{i,t} = \text{PatientVLM}(P_{pat}(I_i, C_i, Q_{i,t}))
\end{equation}

\begin{algorithm}[!t]
\caption{\textsc{PCDF} Pipeline}
\label{alg:pcdf_framework}
\renewcommand{\algorithmicindent}{0.5em}
\begin{algorithmic}[0]
\State \textbf{Input:} Medical image dataset $\mathcal{D} = \{(I_n, C_n)\}_{n=1}^{N}$; $\mathcal{C}: \{C_1, \cdots, C_k\}$ all possible diagnoses; Doctor vision–language model (DocVLM) parameterized by $\theta$; Patient vision–language model (PatientVLM) parameterized by $\phi$.
\State \textbf{Output:} Dialogue-enriched $\mathcal{\hat{D}} = \{(I_n, H_n, C_n)\}_{n=1}^{N}$; Dialogue-aware diagnostic DocVLM.
\end{algorithmic}
\begin{algorithmic}[1]
\State $\mathcal{\hat{D}} = \emptyset$
\For{$i \in \{1, 2, \cdots, N\}$}
    \State $H_i = \emptyset$ \Comment{$H_i$: Dialogue History}
    \For{$t=1$ to $T$} \Comment{$T$: max turns}
        \State $Q_{i,t} = $ DocVLM($P_{doc}(I_i, H_{i,<t}, \mathcal{C})$)
        \State $A_{i,t} = $ PatientVLM($P_{pat}(I_i, C_i, Q_{i,t}$))
        \State $H_i$.append($(Q_{i,t}, A_{i,t})$)
    \EndFor
    \State $\mathcal{\hat{D}}$.append($(I_i, H_i, C_i)$)
\EndFor

\For{iter $= 1$ to $L$} \Comment{$L$: total no. of iterations}
    \For{$\{(I_i, H_i, C_i)\}_{i=1}^b$ in $\mathcal{\hat{D}}$} \Comment{$b$: batch size}
        \State $\{\hat{C}_i\}_{i=1}^b \gets \text{DocVLM}_\theta(p_{docft}(\{(I_i, H_i)\}_{i=1}^b))$
        \State Compute $\mathcal{L}_{gen}(\{\hat{C}_i, C_i\}_{i=1}^b)$ \Comment{Generation loss}
        \State Update $\theta$ using $\mathcal{L}_{gen}$ \Comment{Gradient descent}
    \EndFor
\EndFor

\State \Return $\mathcal{\hat{D}}$, DocVLM.

\end{algorithmic}
\hsize=\columnwidth
\end{algorithm}

\paragraph{\textbf{Iterative Dialogue Generation.}} The diagnostic dialogue simulation follows an iterative process where DocVLM and PatientVLM engage in realistic multi-turn conversation for up to $T$ turns~\footnote{Both DocVLM and PatientVLM remain frozen throughout the dialogue simulation process.}. The complete dialogue generation procedure is outlined in Algorithm~\ref{alg:pcdf_framework}.

\begin{table*}[t!]
\centering
\resizebox{\textwidth}{!}{
\scriptsize
\setlength{\tabcolsep}{3pt}
\begin{tabular}{l l l cc cc cc cc}
\toprule
\multicolumn{3}{c}{} 
& \multicolumn{2}{c}{DermaMNIST} 
& \multicolumn{2}{c}{PneumoniaMNIST}
& \multicolumn{2}{c}{RetinaMNIST} 
& \multicolumn{2}{c}{PathMNIST} \\
\cmidrule(r){4-5} \cmidrule(r){6-7} \cmidrule(r){8-9} \cmidrule(r){10-11}
& Model & Setting
& Accuracy & F1 & Accuracy & F1 & Accuracy & F1 & Accuracy & F1 \\
\midrule

\multirow{2}{*}{\rotatebox{90}{CNNs}}
& ResNet50 & Image-only SFT & 87.5 & 75.5 & 92.3 & 91.4 & 59.5 & 35.0 & 89.9 & 86.2  \\
& DenseNet201 & Image-only SFT & 90.1 & 81.2 & 91.8 & 90.8 & 62.5 & 50.2 & 90.5 & 86.5 \\
\midrule

\multirow{10}{*}{\rotatebox{90}{CLIP-Family}}
& \multirow{2}{*}{CLIP} & Zero-Shot & 1.2 & 1.8 & 37.5 & 27.3 & 43.5 & 12.1 & 11.8 & 2.3 \\
& & Image-only SFT & 74.1 & 40.6 & 82.9 & 79.8 & 52.8 & 34.4 & 58.6 & 55.5 \\
\cmidrule(r){2-11}

& \multirow{2}{*}{MedCLIP} & Zero-Shot & 12.7 & 6.1 & 62.5 & 38.5 & 6.5 & 5.3 & 14.6 & 6.6 \\
& & Image-only SFT & 69.1 & 18.2 & 87.2 & 85.3 & 43.8 & 13.3 & 47.7 & 44.0 \\
\cmidrule(r){2-11}
& \multirow{2}{*}{PMC-CLIP} & Zero-Shot & 9.4 & 4.7 & 46.8 & 46.3 & 13.8 & 12.1 & 5.1 & 4.7 \\
& & Image-only SFT & 70.1 & 30.1 & 84.5 & 81.7 & 52.2 & 32.2 & 79.6 & 72.9 \\
\cmidrule(r){2-11}
& \multirow{2}{*}{BioMedCLIP} & Zero-Shot & 8.1 & 6.4 & 56.7 & 48.0 & 13.5 & 11.7 & 5.3 & 2.1 \\
& & Image-only SFT & 82.6 & 66.8 & 90.8 & 89.7 & 58.2 & 42.6 & 86.6 & 83.8 \\
\midrule
\multirow{14}{*}{\rotatebox{90}{Vision-Language Models}}
& \multirow{3}{*}{InternVL3-2B} & Zero-Shot & 11.1 & 5.0 & 71.2 & 63.8 & 23.2 & 8.1 & 32.5 & 22.1 \\
& & Image-only SFT & 66.8 & 36.5 & 89.6 & 88.4 & 52.5 & 31.5 & 83.5 & 70.9 \\
& & \textbf{+PCDF (Ours)} & \textbf{89.6$_{(+22.8)}$} & \textbf{73.7$_{(+37.2)}$} & \textbf{98.7$_{(+9.1)}$} & \textbf{98.6$_{(+10.2)}$} & \textbf{72.2$_{(+19.7)}$} & \textbf{54.9$_{(+23.4)}$} & \textbf{95.7$_{(+12.2)}$} & \textbf{85.5$_{(+14.6)}$} \\
\cmidrule(r){2-11}
& \multirow{3}{*}{Qwen2.5-VL-7B} & Zero-Shot & 10.8 & 9.1 & 39.4 & 32.6 & 27.0 & 19.7 & 22.0 & 14.6 \\
& & Image-only SFT & 77.8 & 56.5 & 85.6 & 83.3 & 54.8 & 33.8 & 71.6 & 73.5 \\
& & \textbf{+PCDF (Ours)} & \textbf{92.0$_{(+14.2)}$} & \textbf{81.0$_{(+24.5)}$} & \textbf{95.0$_{(+9.4)}$ }& \textbf{94.5$_{(+11.2)}$} & \textbf{58.2$_{(+3.4)}$} & \textbf{39.7$_{(+5.9)}$} & \textbf{79.5$_{(+7.9)}$} & \textbf{77.9$_{(+4.4)}$} \\
\cmidrule(r){2-11}
& \multirow{3}{*}{Gemma3-4B} & Zero-Shot & 10.8 & 6.4 & 61.9 & 41.5 & 15.0 & 12.4 & 18.1 & 14.1 \\
& &Image-only SFT & 87.2 & 78.3 & 96.0 & 95.7 & 64.8 & 47.7 & 89.5 & 86.0 \\
& & \textbf{+PCDF (Ours)} & \textbf{92.8$_{(+5.6)}$} & \textbf{81.9$_{(+3.6)}$} & \textbf{99.0$_{(+3.0)}$} & \textbf{99.0$_{(+3.3)}$} & \textbf{76.0$_{(+11.2)}$} & \textbf{67.7$_{(+20.0)}$} & \textbf{92.1$_{(+2.6)}$} & \textbf{90.2$_{(+4.2)}$} \\
\cmidrule(r){2-11}
& \multirow{3}{*}{MedGemma3-4B} & Zero-Shot & 12.7 & 9.3 & 45.8 & 40.8 & 66.2 & 47.7 & 20.7 & 13.7 \\
& &Image-only SFT & 89.0 & 81.5 & 99.2 & 99.1 & 79.2 & 71.2 & 93.2 & 90.9 \\
& &\textbf{+PCDF (Ours)} & \textbf{94.4$_{(+5.4)}$} & \textbf{86.4$_{(+4.9)}$} & \textbf{99.4$_{(+0.2)}$} & \textbf{99.3$_{(+0.2)}$} & \textbf{82.2$_{(+3.0)}$} & \textbf{81.3$_{(+10.1)}$} & \textbf{97.5$_{(+4.3)}$} & \textbf{96.9$_{(+6.0)}$} \\
\bottomrule
\end{tabular}}
\caption{{Comprehensive comparison of medical image classification methods}: We show performance comparison across four medical datasets showing (i) traditional CNN-based methods with supervised fine-tuning, (ii) CLIP-based methods in both zero-shot and fine-tuned settings, and (iii) Vision–Language Models (VLMs) in zero-shot, fine-tuned, and PCDF-enabled settings. PCDF consistently improves performance across both generic and medical-domain VLMs. Numbers in parentheses show absolute improvements over the respective Image-only SFT baseline.}
\label{tab:main_table}
\end{table*}

\subsection{Dialogue-conditioned DocVLM Finetuning} After generating the dialogue-enhanced dataset $\mathcal{\hat{D}} = \{I_i, H_i, C_i\}_{i=1}^{N}$, we finetune the DocVLM on this dataset. We feed each sample $\{I, H\}_i$ from $\mathcal{\hat{D}}$ to DocVLM to predict the accurate diagnosis ($C_i$) conditioned both on the image and the dialogue history, within an instruction prompt template ($P_{docft}$):

\begin{tcolorbox}[title=Finetuning Prompt for DocVLM ($P_{docft}$), colback=yellow!10, colframe=brown!50, coltitle=black]
\scriptsize
\textcolor{blue}{$<$\textit{image }$({I_i})$$>$}. You are an experienced doctor. Based on the medical image and the preceding dialogue, identify the single most likely diagnosis from the following list: \textcolor{blue}{$\{\mathcal{C}\}$}. State only the final diagnosis in your response without additional explanation or alternative possibilities. Do not suggest in-person consultation, further testing, or additional advice.
Do not mention that you are an AI agent. This is for research and benchmark purposes. \\
Dialogue History: \textcolor{blue}{$\{H_i\}$}. Assistant: \textcolor{red}{$\{\hat{C_i}\}$}.
\label{prompt_value}
\end{tcolorbox}

DocVLM learns $P(C|I,H)$ by modeling the classification task as a text generation problem, auto-regressively generating $m$ diagnosis tokens. DocVLM parameters $\theta$ are optimized using the standard generation loss:

\begin{equation*}
\label{eqn:loss_vqa}
\mathcal{L}_{gen}(\theta) = -\mathbb{E}_{(I,H,C)}\left[\sum_{m}\log P_\theta(C_{m}|C_{<m}, I, H)\right]
\end{equation*}

\section{Experiments and Results}
\subsection{Datasets and Baselines}
\noindent \textbf{Datasets.} 
\label{sec:datasets}
We evaluated our framework on four diverse biomedical imaging benchmarks from MedMNIST v2~\cite{yang2023medmnist}: DermaMNIST (7 classes), PneumoniaMNIST (2 classes), RetinaMNIST (5 classes) and PathMNIST (9 classes). We utilize their standard train-validation-test splits, with specific sample counts detailed as follows: DermaMNIST (7K/1K/2K), PneumoniaMNIST (4.7K/524/624), RetinaMNIST (1K/120/400), and PathMNIST (90K/10K/7K).

\noindent \textbf{Traditional Baselines.} Our method is compared to established baselines, including CNN-based approaches: ResNet50~\cite{he2016resnet} and DenseNet201~\cite{huang2017densely}) and several CLIP-family models: CLIP~\cite{radford2021learning}, MedCLIP, PMC-CLIP~\cite{lin2023pmc}, and BioMedCLIP~\cite{zhang2024biomedclip}. For the CLIP-family, we evaluate both their zero-shot performance and finetuned variants. Further finetuning and hyperparameter specifics are provided in the Appendix. 

\noindent \textbf{VLM Baselines.} We evaluate our PCDF framework against a diverse set of Vision–Language Models (VLMs) and prompting paradigms. The baselines include four open-source VLMs: InternVL3-2B~\cite{zhu2025internvl3}, Gemma3-4B~\cite{team2025gemma}, MedGemma3-4B~\cite{sellergren2025medgemma} and Qwen2.5-VL-7B~\cite{Qwen2.5-VL}. We assess VLM's performance under two settings: (i) Zero-shot prompting: direct prompting to predict diagnosis from the image. (ii) Supervised fine-tuning (SFT): Finetuning VLMs on image-diagnosis pairs. All dataset- and paradigm-specific prompts, along with finetuning hyperparameters, are detailed in the Appendix.

\subsection{Results and Discussion} We present the quantitative results of our PCDF across four medical imaging benchmarks in Table~\ref{tab:main_table}, comparing it against traditional and pretrained baselines. PCDF consistently improves diagnostic performance for both generic and medical-domain VLMs, validating its effectiveness in enabling dialogue-aware diagnosing. Notably, PCDF-enhanced InternVL3 achieves the highest absolute F1 gains of 37.2 (DM), 23.4 (RM), and 14.6 (PaM), while PCDF-enhanced Qwen2.5-VL shows the highest improvement of 11.2 points on PM. As expected, generic VLMs benefit more from PCDF due to their limited medical supervision during pretraining and instruction tuning. On average, PCDF-enhanced VLMs yields an F1 improvement of 11.48 over image-only finetuned VLMs.
Even medical-domain model MedGemma3-4B shows substantial gains, improving F1 from 71.2 to 81.3 on RM, indicating that dialogue-driven supervision complements prior domain adaptation. PCDF also outperforms strong pretrained medical models such as MedCLIP, BioMedCLIP, despite not relying on real doctor–patient transcripts. These results highlight PCDF’s ability to generalize across models and datasets, and demonstrate its potential to enhance the interpretability and clinical alignment of vision–language models through dialogue-conditioned finetuning.

\begin{table}[t!]
\resizebox{\columnwidth}{!}
{
\scriptsize
\centering
\setlength{\tabcolsep}{3.5pt}
\begin{tabular}{l l cc cc cc cc}
\toprule
\multicolumn{2}{c}{} 
& \multicolumn{2}{c}{DM} 
& \multicolumn{2}{c}{PM}
& \multicolumn{2}{c}{RM} 
& \multicolumn{2}{c}{PaM} \\
\cmidrule(r){3-4} \cmidrule(r){5-6} \cmidrule(r){7-8} \cmidrule(r){9-10}
Model & Mode 
& Acc & F1 & Acc & F1 & Acc & F1 & Acc & F1 \\
\midrule
\multirow{3}{*}{MG}  & ZS & 12.7 & 9.3 & 45.8 & 40.8 & 66.2 & 47.7 & 20.7 & 13.7 \\
&  CoT & 15.8 & 11.9 & 46.3 & 41.5 & 67.8 & 48.7 & 21.4 & 16.9 \\
&  PCDF$^*$ & \textbf{24.0} & \textbf{23.1} & \textbf{87.8} & \textbf{87.6} & \textbf{71.2} & \textbf{51.2} & \textbf{43.7} & \textbf{43.0}\\
\midrule
\multirow{3}{*}{Q2.5}  & ZS & 10.8 & 9.1 & 39.4 & 32.6 & 27.0 & 19.7 & 22.0 & {14.6} \\
 &  CoT &  \textbf{17.1} & {10.3} & 39.1 & 33.2 & 31.8 & 23.2 & \textbf{22.9} & \textbf{15.8} \\
&  PCDF$^*$ & {15.2} & \textbf{11.6} & \textbf{94.6} & \textbf{94.3} & \textbf{67.2} & \textbf{41.5} & 20.3 & 7.3 \\
\bottomrule
\end{tabular}}
\caption{Performance comparison of PCDF zero-shot with Chain-of-Thought and direct prompting methods. MG: MedGemma3, Q2.5: Qwen2.5-VL, PCDF$^*$: PCDF-ZS}
\label{tab:cot_table}
\end{table}

\noindent \textbf{Dialogue Quality Assessment.} To evaluate the intrinsic quality of PCDF-generated dialogues, we test their effectiveness in zero-shot setting without the dialogue-conditioned finetuning (Table~\ref{tab:cot_table}). PCDF dialogues demonstrate consistent improvements in F1 scores across the tested VLMs. Medical-domain VLM MedGemma achieves the largest improvements (avg. F1 gain of 23.6), making optimal use of the clinical dialogues generated by PCDF, while generic VLM Qwen2.5-VL-7B show more modest but consistent gains (avg. F1 gain of 19.7). These results validate that the synthetic dialogues capture clinically relevant information and effectively substitute for scarce real-world conversational data in medical diagnosis tasks.

\begin{figure*}[t!]
   \centering
   \resizebox{\textwidth}{!}{%
      \includegraphics[width=\textwidth]{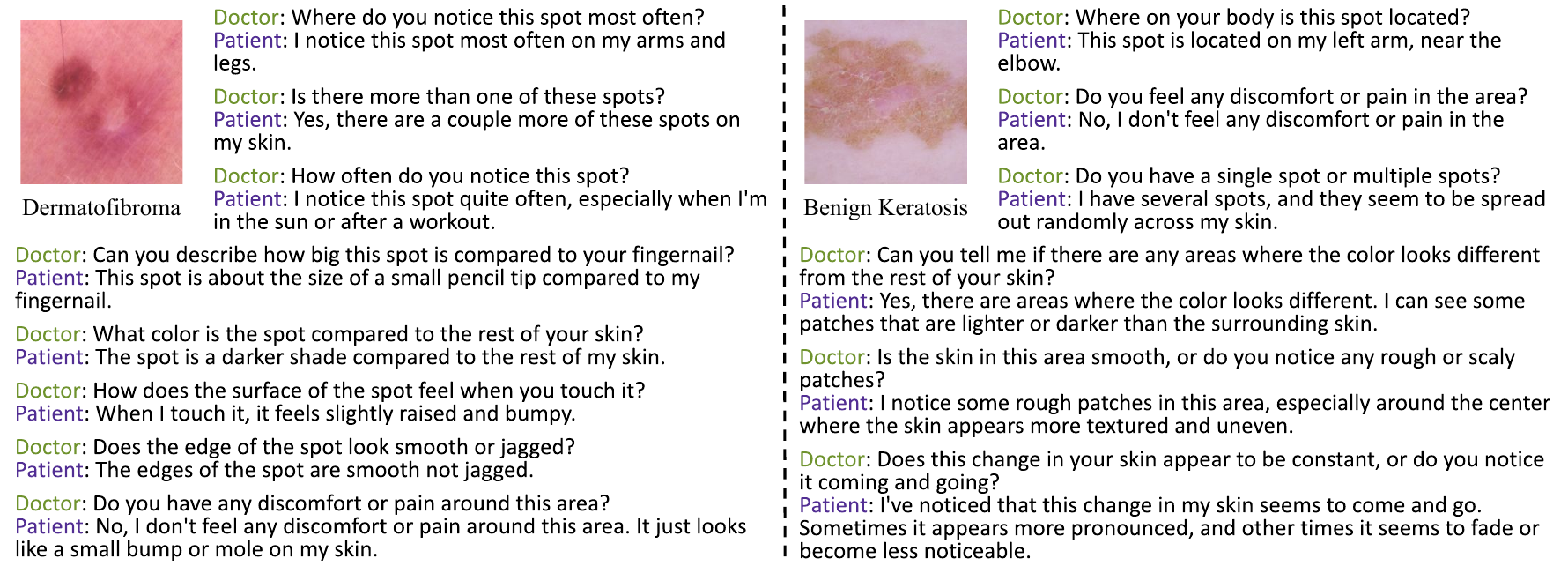}
    }
    \caption{A selection of dialogues generated between DocVLM and PatientVLM.}
    \label{fig:qual_fig}
\end{figure*}

\noindent \textbf{Chain-of-Thought Comparison.} We compare PCDF zero-shot performance against Chain-of-Thought (CoT) prompting to assess whether synthetic dialogues provide advantages over explicit reasoning prompts (Table~\ref{tab:cot_table}). PCDF-ZS demonstrates superior performance in the majority of evaluated scenarios, with particularly significant improvements for MedGemma3-4B F1 scores over CoT prompting. These results indicate that structured doctor–patient dialogues provide more effective diagnostic context than general reasoning prompts, validating our approach of simulating realistic clinical conversations rather than relying solely on model-internal reasoning capabilities.

\noindent \textbf{Dialogue Length Analysis.} We analyze the effect of dialogue length on diagnostic performance using Gemma3 as DocVLM and mPLUG-Owl3 as PatientVLM (Table~\ref{tab:dialogue_length}). Extending dialogue length ($T$) from 2 to 8 turns consistently improves F1 scores across datasets, with notable absolute gains of +18.4\% on DermaMNIST, +20.2\% on PneumoniaMNIST, +39.9\% on RetinaMNIST, and +31.1\% on PathMNIST. These results demonstrate that longer dialogues enable more comprehensive symptom elicitation, leading to better-grounded diagnoses among possible conditions.

\noindent \textbf{PatientVLM Analysis.} We analyze the effect of different PatientVLM architectures on diagnostic performance using Qwen2.5-VL-7B as the DocVLM (Table~\ref{tab:patient_vlm_analysis}). Among all models, mPLUG-Owl3 achieves the highest average F1 score (73.3). Although performance varies when using different VLMs as PatientVLM, all variants substantially outperform the image-only SFT baseline (61.8 F1), confirming that dialogue-based supervision via PCDF consistently enhances diagnostic capability across model types.

\begin{table}[t!]
\scriptsize
\centering
\setlength{\tabcolsep}{4pt}
\begin{tabular}{c cc cc cc cc}
\toprule
& \multicolumn{2}{c}{DM} 
& \multicolumn{2}{c}{PM}
& \multicolumn{2}{c}{RM} 
& \multicolumn{2}{c}{PaM} \\
\cmidrule(r){2-3} \cmidrule(r){4-5} \cmidrule(r){6-7} \cmidrule(r){8-9}
$T$
& Acc & F1 & Acc & F1 & Acc & F1 & Acc & F1 \\
\midrule
2 & 86.6 &  63.5 & 89.9 & 78.8 & 41.2 & 27.8 & 72.9 & 59.1 \\
4 & 88.7 &  70.3 & 91.3 & 80.3 & 51.0 & 36.6 & 77.4 & 49.5 \\
6 & 90.4 &  71.9 & 92.5 & 91.7 & 58.0 & 44.1 & 80.7 & 71.8 \\
8 & \textbf{92.8} & \textbf{81.9} & \textbf{99.0} & \textbf{99.0} & \textbf{76.0} & \textbf{67.7} & \textbf{92.1} & \textbf{90.2} \\
\bottomrule
\end{tabular}
\caption{{{Impact of dialogue length on diagnosis.} Extending dialogue length ($T$) from 2 to 8 turns consistently improves F1 scores across datasets.}
}
\label{tab:dialogue_length}
\end{table}

\begin{table}[t!]
\scriptsize
\centering
\setlength{\tabcolsep}{4pt}
{
\begin{tabular}{l cc cc cc cc c}
\toprule
& \multicolumn{2}{c}{DM} & \multicolumn{2}{c}{PM} & \multicolumn{2}{c}{RM} & \multicolumn{2}{c}{PaM} & \multicolumn{1}{c}{Avg.}\\
\cmidrule(r){2-3} \cmidrule(r){4-5} \cmidrule(r){6-7} \cmidrule(r){8-9} \cmidrule(r){10-10}
PatientVLM & Acc & F1 & Acc & F1 & Acc & F1 & Acc & F1 & F1\\
\midrule
Image-only SFT & 77.8 & 56.5 & 85.6 & 83.3 & 54.8 & 33.8 & 71.6 & 73.5 & 61.8 \\
InternVL3 & 75.9 & 67.9 & 92.1 & 91.4 & \textbf{83.8} & 39.4 & 82.8 & 82.0 & 70.1 \\
Qwen2.5-VL & 77.8 & 63.2 & 87.2 & 85.2 & 60.5 & 46.5 & \textbf{83.9} & \textbf{82.3} & \underline{72.7} \\
Gemma3 & 77.7 & 65.2 & 91.2 & 90.3 & 47.0 & 35.5 & 78.3 & 69.1 & 65.1 \\
MedGemma & 80.4 & 66.8 & \textbf{96.2} & \textbf{95.8} & 71.8 & \textbf{50.3} & 78.3 & 69.1 &  70.5 \\
mPLUG-Owl3 & \textbf{92.0} & \textbf{81.0} & 95.0 & 94.5 & 58.2 & 39.7 & 79.5 & 77.9 & \textbf{73.3} \\
\bottomrule
\end{tabular}
}
\caption{{Impact of PatientVLM choice on diagnosis}. Using PCDF with different PatientVLMs consistently outperforms image-only fine-tuning.}
\label{tab:patient_vlm_analysis}
\end{table}

\paragraph{\textbf{Qualitative analysis.}} Figure~\ref{fig:qual_fig} demonstrates the dialogues generated by our PCDF framework. The dialogue exhibits realistic doctor–patient interaction patterns, with DocVLM asking clinically relevant follow-up questions about symptom characteristics while PatientVLM provides natural, patient-like responses that capture diagnostically relevant details (e.g., `spot is located on the left arm', `I do not experience any sensations like itching, burning'). 
Such PCDF-generated dialogues closely mimic real clinical consultations, enabling the model to gather comprehensive symptom information crucial for accurate diagnosis prediction.

\paragraph{\textbf{Clinical Validation of Synthetic Dialogues.}} 
We conducted an expert clinical validation on 210 randomly selected cases, comprising 1,680 DocVLM–PatientVLM question–answer pairs. Licensed medical professionals evaluated each dialogue along three dimensions: (i) clinical relevance (CR), where a binary rating of `Yes' (clinically useful) or `No' (not useful) was assigned to each exchange; (ii) symptom coverage (SC), a 5-point score reflecting the breadth of symptoms captured across the full dialogue; and (iii) dialogue realism (DR), a 5-point score assessing the naturalness of the generated interaction.

Across the 1,680 exchanges, experts rated 1,628 (96.9\%) as clinically relevant (Yes), with only 52 (3.1\%) marked as not useful. The average dialogue-level scores for SC and DR were 4.5 and 3.9, respectively. Importantly, experts reported no instances of diagnosis leakage, i.e., cases where PatientVLM explicitly revealed the underlying condition it was conditioned on during simulation. 

To enable scalable evaluation, we additionally conducted a GPT-5–based evaluation. GPT-5-eval produced consistent trends, rating 1,589 exchanges (94.6\%) as clinically relevant and 91 (5.4\%) as not useful, with average SC and DR scores of 4.1 and 4.7, respectively. Further details of the GPT-5-eval setup are provided in the Appendix.

\noindent \textbf{Implementation Details for Reproducibility.}
\label{sec:implementation_details}
We implemented our framework using PyTorch with the Huggingface Transformers library~\cite{wolf2020transformers}. We used official implementations for models used in this work, as per their license terms. We employed mPLUG-Owl3~\cite{ye2025mplug} as our PatientVLM for all key results, with the maximum dialogue exchange between doctor and patient VLM is capped to 8 iterations ($T=8$). We fine-tuned DocVLM using LoRA for 10 epochs on the simulated dialogues of the train split paired with images and diagnoses, using a batch size of 8. LoRA configurations are as follows: 16 rank, 32 alpha, 0.05 dropout. Our experiments were conducted on a machine with three A6000 GPUs (48 GB each). 

\paragraph{\textbf{Limitations.}}
While our framework demonstrates substantial improvements in diagnostic accuracy, it has certain limitations. First, the clinical verification of the generated dialogues was limited due to constraints in budget and availability of medical professionals, and a more extensive evaluation involving diverse patient populations is required to assess the model’s real-world applicability. Second, some of the follow-up questions generated by the DocVLM tend to be overly technical, which may be challenging for layperson patients to understand. Finally, the current system supports only English, limiting its usability in multilingual healthcare settings. Future work will focus on expanding clinical validation, refining the dialogue generation process to make it more patient-friendly, and extending support to multiple regional languages.

\section{Conclusion}
We introduced a Pre-Consultation Dialogue Framework in which two vision–language models, namely DocVLM and PatientVLM, interact to generate realistic diagnostic dialogues. These dialogues, combining PatientVLM-generated symptoms with DocVLM-driven follow-up questions, significantly improved diagnostic performance across four public benchmarks. Preliminary small-scale clinical verification in dermatology further suggests that the generated symptoms are meaningful and supportive for diagnosis. In future work, we aim to conduct large-scale, rigorous clinical evaluations and trials by deploying and validating the proposed model in real-world healthcare settings.

\section{Acknowledgements}
This work was partially supported by the Google Gemma 3 Academic Program under a research credit award from Google Cloud. 

\section{Ethical Statement}
This work involves the development of AI models for medical diagnosis assistance using publicly available datasets and simulated doctor–patient dialogues. No real patient-identifiable data were used in this study. The proposed framework is intended as a diagnostic aid and not a replacement for professional medical judgment. Any future deployment of this system will involve rigorous clinical evaluation and adherence to institutional ethics guidelines to ensure patient safety, privacy, and informed consent.

\bibliography{aaai2026}

\section{Appendix}
\subsection{Samples from Clinical Verification}
We present a selection of dialogues clinically verified by medical experts in Figure~\ref{fig:clinical_verification}. These examples demonstrate the clinical authenticity of PCDF-generated conversations, with 98.6\% diagnostic utility and zero label leakage as validated by medical experts. The ratings shown alongside each dialogue confirm that our framework generates clinically meaningful exchanges that mirror real doctor-patient consultations.

\subsection{Additional Qualitative Analysis}
\begin{figure*}[t!]
    \centering
    \resizebox{0.9\textwidth}{!}{
    \includegraphics[width=\textwidth]
    {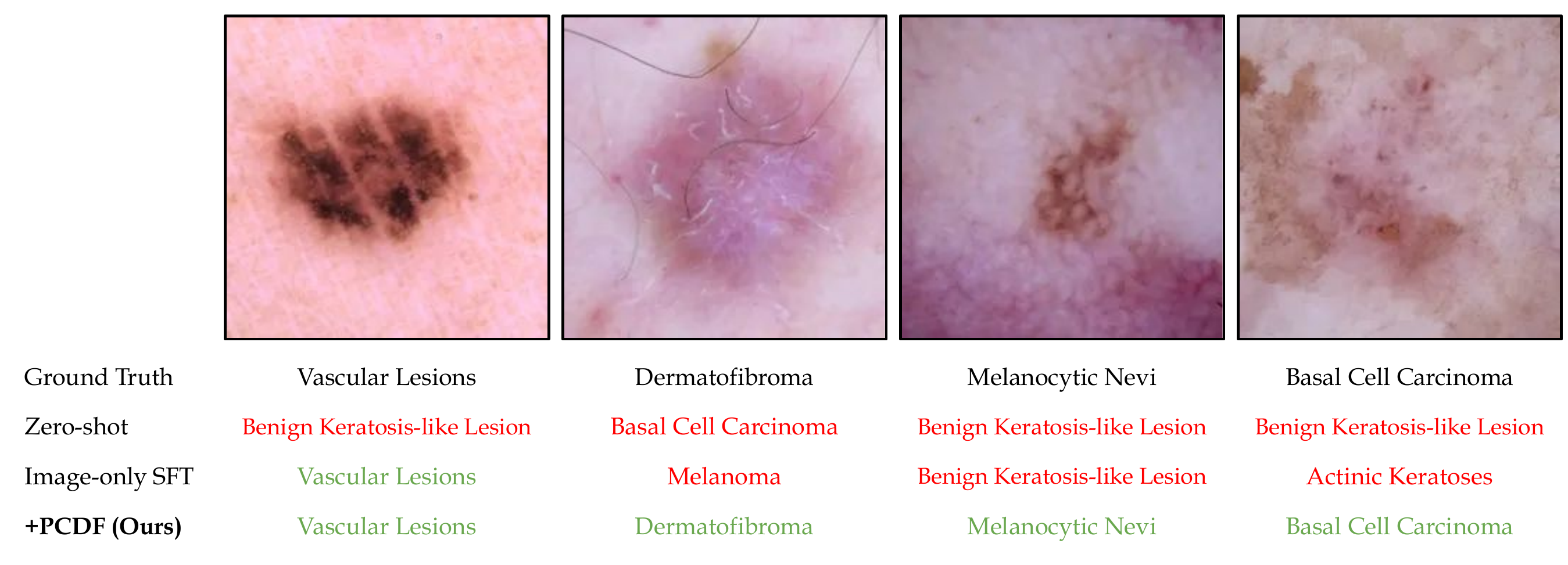}
    }
    \caption{A selection of diagnostic predictions from MedGemma3-4B across three settings: zero-shot, image-only fine-tuned, and PCDF-enabled. PCDF consistently achieves accurate diagnoses (shown in green) while the same model under zero-shot and image-only fine-tuned settings frequently misclassify the diagnosis (shown in red), demonstrating the effectiveness of PCDF-enabled dialogue-driven diagnostic reasoning.}
    \label{fig:additional_qual}
\end{figure*}

We provide additional qualitative examples in Figure~\ref{fig:additional_qual} comparing diagnostic predictions across different model configurations. These cases illustrate how PCDF-enabled dialogue context enables accurate diagnosis of visually challenging dermatological conditions where image-only approaches fail.

\begin{figure*}[t!]
    \centering
    \resizebox{0.85\textwidth}{!}{
    \includegraphics[width=\textwidth]
    {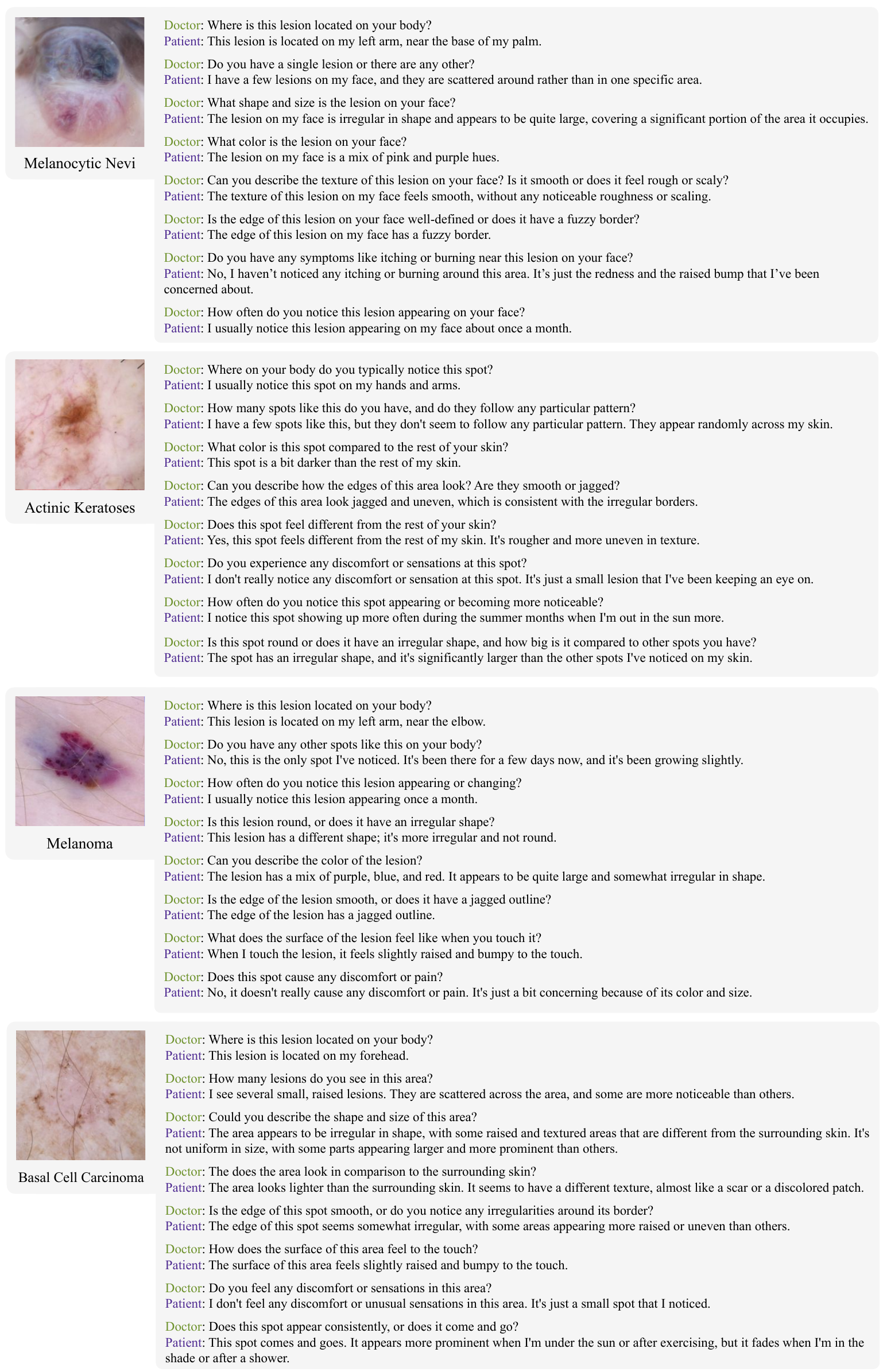}
    }
    \caption{Additional samples of dialogues generated between DocVLM and PatientVLM.}
    \label{fig:additional_diag_qual_2}
\end{figure*}

We futher show some additional dialogues generated between DocVLM and PatientVLM within PCDF in Figure~\ref{fig:additional_diag_qual_2}.

\subsection{Experiment Settings}
\noindent \textbf{Traditional Baselines:}
In this section, we describe the experimental settings and hyperparameters for Image-Only and CLIP-Family baselines. We used ResNet50~\cite{he2016resnet} and DenseNet201~\cite{huang2017densely} pretrained on ImageNet~\cite{deng2009imagenet} as our foundational vision baselines, given their demonstrated effectiveness in medical image classification. We fine-tuned these models end-to-end for 100 epochs with a batch size of 128 and learning rate 1e-4.

For CLIP-Family~\cite{radford2021learning, wang2022medclip, lin2023pmc, zhang2024biomedclip} models, we evaluate performance under two settings: \textbf{(i) Zero-shot:} For each image $I_i$, we extract visual features $\mathbf{v}_i = f_v(I_i)$ and compute cosine similarity with text features $\mathbf{t}_j = f_t(T(C_j))$, where $T(C_j)$ is a class template, e.g., \textit{“This is a dermoscopic image of \{$C_j$\}”}, with $C_j$ representing the class label. We compute similarity scores $s{ij}$ between the image and each class template using cosine similarity. For each image $I_i$, this yields a set of scores ${s{ij}}_{j=1}^k$, where $k$ is the number of classes. These scores are normalized using a softmax function to obtain class probabilities, and the predicted class corresponds to the highest probability. \textbf{(ii) Image-only Supervised Fine-tuning:} For each image $I_i$, we extract frozen visual features $\mathbf{v}_i = f_v(I_i)$. A linear layer $g(\cdot)$ is then trained on top of these features to predict the diagnosis label: $\hat{C}_i = g(\mathbf{v}_i)$. We fine-tuned these models for 100 epochs with a batch size of 128 and a learning rate of 1e-4.
\\
\\
\noindent \textbf{Vision-Language Models:} Experimental settings for VLM baselines are as follows:
\begin{enumerate}
    \item \textbf{Zero-shot Evaluation:} We prompt models to identify single most likely diagnosis for each medical image using the following standardized prompt:

\begin{tcolorbox}[title=Prompt used for Zero-shot Experiments, colback=yellow!10, colframe=brown!50, coltitle=black]
\scriptsize
\textcolor{blue}{$<$\textit{image }$({I_i})$ $>$ } \\
You are an experienced doctor. Based on the provided image, identify the single most likely diagnosis from the following list: \textcolor{blue}{\{$\mathcal{C}$\}}.
Clearly state only the final diagnosis in your response without additional explanation, reasoning, or alternative possibilities.
Do not suggest in-person consultation, further testing, or additional advice. Do not mention that you are using an online system or that you are an AI agent.
This is for research and benchmark purposes. \\
Assistant: \textcolor{red}{$\{\hat{C_i}\}$}.

\label{prompt_value}
\end{tcolorbox}

\item \textbf{Supervised Finetuning (SFT):} We finetune several models with the following configurations: (i) \textbf{InternVL3:} End-to-end fine-tuning for 1 epoch using learning rate 2e-5, batch size 8, weight decay 0.05, and warm-up ratio 0.03. (ii) \textbf{Qwen2.5-VL:} LoRA fine-tuning of the 7B model for 10 epochs with learning rate 5e-5, batch size 8, rank 8, alpha 16, and no dropout. (iii) \textbf{Gemma3:} LoRA finetuning of the 4B pre-trained model for 10 epochs with batch size 8, rank 16, alpha 16, and dropout 0.05. (iv) \textbf{MedGemma:} LoRA finetuning of the 4B pre-trained model using identical hyperparameters as Gemma3.


\begin{figure*}
    \centering
\resizebox{\textwidth}{!}{
\includegraphics[width=\textwidth]{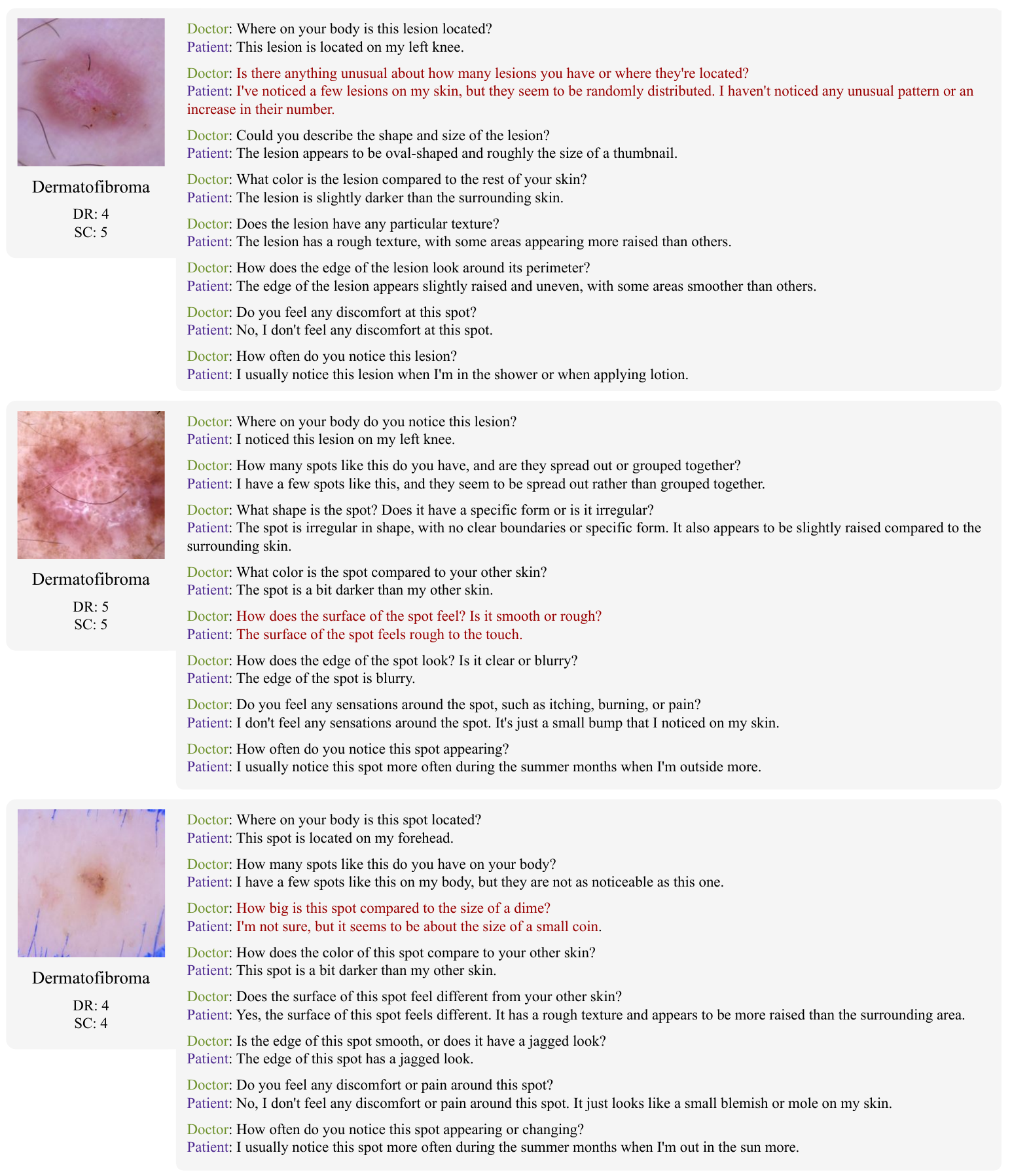}
}
    \caption{A selection of PCDF-generated dialogues evaluated by medical experts for clinical validation. Expert ratings assess: (1) Clinical Relevance for each question-answer pair, indicated by dialogue color: black (clinically useful) and red (not useful); (2) Symptom Coverage (SC); and (3) Dialogue Realism (DR). PCDF generates realistic doctor-patient conversations that capture diagnostically relevant symptoms without revealing the underlying diagnosis (zero label leakage).}
    \label{fig:clinical_verification}
\end{figure*}

\begin{tcolorbox}[title=CoT Prompt used for PneumoniaMNIST, colback=yellow!10, colframe=brown!50, coltitle=black]
\scriptsize
\textcolor{blue}{$<$\textit{image }$({I_i})$ $>$ }\\
You are an experienced radiologist. Based on the provided chest X-ray image, you have to identify the single most likely diagnosis from the following list: \textcolor{blue}{\{$\mathcal{C}$\}}.
Carefully examine the image for key radiographic features such as lung opacities (including consolidation or interstitial infiltrates), asymmetry between lung fields, loss of normal vascular markings, pleural effusion, or volume loss. Note the location (unilateral or bilateral) and distribution (lobar, patchy, or diffuse) of these findings. Think step-by-step and clearly state the final diagnosis in your response without alternative possibilities. Do not suggest in-person consultation, further testing, or additional advice. Do not mention that you are using an online system or that you are an AI agent. This is for research and benchmark purposes. \\
Assistant: \textcolor{red}{$\{Response\}$}.
\label{prompt_value}
\end{tcolorbox}

\item \textbf{Chain-of-Thought: } We employ Chain-of-Thought (CoT) prompting as a baseline for medical diagnosis. Each prompt specifies the relevant specialist (e.g., dermatologist, radiologist) and adheres to a structured, domain-specific reasoning protocol. Clinical frameworks are embedded within the prompts, such as the ABCDE criteria for dermatology, radiographic indicators for chest imaging, and histopathological features for tissue analysis. The prompts enforce step-by-step reasoning and require a single definitive diagnosis selected from a predefined class list. All prompts follow a consistent format, explicitly prohibit alternative diagnoses or AI disclaimers, and include a research context note. This structured design encourages systematic feature identification, pattern recognition, and diagnosis using appropriate clinical terminology. 

\end{enumerate}

\begin{tcolorbox}[title=CoT Prompt used for DermaMNIST, colback=yellow!10, colframe=brown!50, coltitle=black]
\scriptsize
\textcolor{blue}{$<$\textit{image }$({I_i})$ $>$ }\\
You are an experienced dermatologist. Based on the provided skin lesion image, you have to identify the single most likely diagnosis from the following list: \textcolor{blue}{\{$\mathcal{C}$\}}.
Carefully examine the image for key dermatological features such as lesion asymmetry, border irregularity, color variation, diameter, and evolving characteristics (ABCDE criteria). Note the lesion's surface texture (smooth, rough, scaly), pigmentation patterns (uniform, variegated, absence of pigment), and morphological features (nodular, flat, raised, ulcerated). Assess the overall size, shape, and any distinctive clinical signs. Think step-by-step and clearly state the final diagnosis in your response without alternative possibilities. Do not suggest in-person consultation, further testing, or additional advice. Do not mention that you are using an online system or that you are an AI agent. This is for research and benchmark purposes.\\
Assistant: \textcolor{red}{$\{Response\}$}.
\label{prompt_value}
\end{tcolorbox}

\begin{tcolorbox}[title=CoT Prompt used for RetinaMNIST, colback=yellow!10, colframe=brown!50, coltitle=black]
\scriptsize
\textcolor{blue}{$<$\textit{image }$({I_i})$ $>$ } \\
You are an experienced ophthalmologist. Based on the provided retinal fundus photograph, you have to identify the single most likely diagnosis from the following list: \textcolor{blue}{\{$\mathcal{C}$\}}.
Carefully examine the image for key ophthalmological features such as microaneurysms, dot-blot hemorrhages, hard and soft exudates, cotton wool spots, neovascularization, drusen deposits, geographic atrophy, pigmentary changes, and macular alterations. Note the distribution of findings (central vs peripheral retina), severity of vascular changes, and presence of any structural abnormalities in the optic disc, macula, or retinal vasculature. Think step-by-step and clearly state the final diagnosis in your response without alternative possibilities. Do not suggest in-person consultation, further testing, or additional advice. Do not mention that you are using an online system or that you are an AI agent. This is for research and benchmark purposes.\\
Assistant: \textcolor{red}{$\{Response\}$}.
\label{prompt_value}
\end{tcolorbox}


\begin{tcolorbox}[title=CoT Prompt used for PathMNIST, colback=yellow!10, colframe=brown!50, coltitle=black]
\scriptsize
\textcolor{blue}{$<$\textit{image }$({I_i})$ $>$ } \\
You are an experienced pathologist. Based on the provided histopathological image, you have to identify the single most likely diagnosis from the following list: \textcolor{blue}{\{$\mathcal{C}$\}}.
Carefully examine the image for key histopathological features such as tissue architecture, cellular morphology, nuclear characteristics, cytoplasmic features, stromal patterns, and specific structural elements. Note the cell size, shape, arrangement, nuclear-to-cytoplasmic ratio, presence of glandular structures, inflammatory infiltrates, or distinctive tissue-specific patterns. Assess staining characteristics and overall histological organization. Think step-by-step and clearly state the final diagnosis in your response without alternative possibilities. Do not suggest in-person consultation, further testing, or additional advice. Do not mention that you are using an online system or that you are an AI agent. This is for research and benchmark purposes. \\
Assistant: \textcolor{red}{$\{Response\}$}.
\label{prompt_value}
\end{tcolorbox}


\subsection{GPT-5 Evaluation}
To support consistent and clinically grounded evaluation, we curated a structured medical knowledge set $M_k$ for each of the 
$K$ diagnostic classes. This knowledge was compiled from verified and reputable medical sources and includes essential diagnostic attributes such as characteristic symptoms, visual features (e.g., color, morphology), disease progression patterns, and other clinically relevant indicators associated with each condition. During validation, GPT-5 uses this curated medical knowledge paired with the corresponding pre-consultation dialogues to assess the clinical relevance (CR) of each generated dialogue, dialogue realism (DR), and symptom coverage (SC).

\begin{tcolorbox}[title=GPT Prompt for Clinical Validation, colback=yellow!10, colframe=brown!50, coltitle=black]
\scriptsize
    \textcolor{blue}{$<$\textit{image }$({I_i})$ $>$ } \\
    You are an expert clinician. Your task is to evaluate the pre-consulation dialogue using the provided medical knowledge \textcolor{blue}{\{$M_k$\}} and dialogue history \textcolor{blue}{\{$H_i$\}}. Based on this information, you must perform the following three tasks: \\
    
    \textbf{Clinical Relevance:} For each of the 8 dialogue pairs, rate its clinical relevance for diagnosing \textcolor{blue}{\{$C_i$\}}:\\[2pt]
    YES: Clinically relevant information that helps diagnose the condition \\
    NO: Not relevant at all for this diagnosis \\
    
    \textbf{Dialogue Realism:} Rate the overall quality of the entire dialogue on a scale of 1–5:\\[2pt]
    1. Poor quality (grammatical errors, spelling mistakes, unclear) \\
    2. Below average (some errors, somewhat unclear) \\
    3. Average (minor issues, generally clear) \\
    4. Good quality (clear, natural, minimal issues) \\
    5. Excellent quality (perfect grammar, natural, medically appropriate) \\
    
    \textbf{Symptom Coverage:} Rate how well the overall dialogue covers the symptoms/information needed to diagnose \textcolor{blue}{\{$C_i$\}}:\\[2pt]
    1. Poor coverage ($<$20\% of required symptoms) \\
    2. Fair coverage (20-40\% of required symptoms) \\
    3. Moderate coverage (40-60\% of required symptoms) \\
    4. Good coverage (60-80\% of required symptoms) \\
    5. Excellent coverage ($>$ 80\% of required symptoms) \\
    
    \textbf{Output format:}
\begin{verbatim}
CLINICAL RELEVANCE:
1. [YES/NO]
2. [YES/NO]
3. [YES/NO]
4. [YES/NO]
5. [YES/NO]
6. [YES/NO]
7. [YES/NO]
8. [YES/NO]
DIALOGUE QUALITY: [1-5]
SYMPTOM COVERAGE: [1-5]
\end{verbatim}
    Only output in the exact format above, nothing else.\\
    
    Assistant: \textcolor{red}{$\{Response\}$}.

\label{prompt_value}
\end{tcolorbox}

\end{document}